%% file: main.tex
\documentclass[journal]{IEEEtran}
\usepackage{cite}
\usepackage{enumerate}
\usepackage[english]{babel}
\usepackage[utf8]{inputenc}
\usepackage{algorithm}
\usepackage[noend]{algpseudocode}
\usepackage{amsmath}
\usepackage{amssymb}
\usepackage{amsthm}
\usepackage{graphicx}
\usepackage{hyperref}
\usepackage{graphics}
\usepackage[dvipsnames]{xcolor}
\usepackage{caption}
\usepackage{subcaption}
\usepackage{multirow}

\newtheorem{lemma}{Lemma}
\newtheorem{theorem}{Theorem}
\def\proof{\noindent\hspace{2em}{\itshape Proof: }}

\newtheorem{remark}{Remark}

\newcommand{\R}{\mathbb{R}}
\newcommand{\E}{\mathbb{E}}

\newcommand{\g}{\gamma}

\usepackage{tikz}
\usepackage{textcomp}
\usepackage{hyperref}
\usepackage{lipsum}

\newcommand\copyrighttext{%
  \footnotesize \textcopyright 2021 IEEE. Personal use is permitted, but republication/redistribution requires IEEE permission. This paper is accepted for publication at IEEE Transactions on Automatic Control.
  DOI: 10.1109/TAC.2021.3112851}
\newcommand\copyrightnotice{%
\begin{tikzpicture}[remember picture,overlay]
\node[anchor=south,yshift=10pt] at (current page.south) {\fbox{\parbox{\dimexpr\textwidth-\fboxsep-\fboxrule\relax}{\copyrighttext}}};
\end{tikzpicture}%
}

\IEEEoverridecommandlockouts    
\title{Generalized Second Order Value Iteration in Markov Decision Processes}

\author{Chandramouli Kamanchi$^{*}$, Raghuram Bharadwaj Diddigi$^{*}$, Shalabh Bhatnagar
\thanks{$^{*}$ Equal Contribution.}%
\thanks{The authors are with the Department of Computer Science and Automation, Indian Institute of Science (IISc), Bengaluru 560012,
India. (E-mails: \{chandramouli, raghub, shalabh\}@iisc.ac.in).}%
\thanks{Raghuram Bharadwaj was supported by a fellowship grant from the Centre for Networked Intelligence (a Cisco CSR initiative) of the Indian Institute of Science, Bangalore. Shalabh Bhatnagar was supported by the J.C.Bose Fellowship, a project from DST under the ICPS Program and the RBCCPS, IISc.}%
}
\begin{document}

\maketitle
\thispagestyle{empty}
\pagestyle{empty}
\copyrightnotice

\begin{abstract}
 Value iteration is a fixed point iteration technique utilized to obtain the optimal value function and policy in a discounted reward Markov Decision Process (MDP). Here, a contraction operator is constructed and applied repeatedly to arrive at the optimal solution. Value iteration is a first order method and therefore it may take a large number of iterations to converge to the optimal solution. Successive relaxation is a popular technique that can be applied to solve a fixed point equation. It has been shown in the literature that, under a special structure of the MDP, successive over-relaxation technique computes the optimal value function faster than standard value iteration. In this work, we propose a second order value iteration procedure that is obtained by applying the Newton-Raphson method to the successive relaxation value iteration scheme. We prove the global convergence of our algorithm to the optimal solution asymptotically and show the second order convergence. Through experiments, we demonstrate the effectiveness of our proposed approach.  
\end{abstract}

\input{introduction}

\input{related-work}
\input{background}
\input{proposed-algorithm}
\input{convergence-analysis}

\input{experiments}

\input{conclusion}

\bibliographystyle{plain}
\bibliography{references}

\end{document}

%% file: introduction.tex
\section{Introduction}
In a discounted reward Markov Decision Process \cite{bertsekas1996neuro}, the objective is to maximize the expected cumulative discounted reward. Reinforcement Learning (RL) deals with the algorithms for solving an MDP problem when the model information (i.e., probability transition matrix and reward function) is unknown. RL algorithms instead make use of state and reward samples and estimate the optimal value function and policy. Due to the success of deep learning \cite{lecun2015deep}, RL algorithms in combination with deep neural networks have been successfully deployed to solve many real world problems and games \cite{mnih2013playing}. However, there is ongoing research for improving the sample-efficiency as well as convergence of RL algorithms \cite{haarnoja2018soft}. 

Many RL algorithms can be viewed as stochastic approximation \cite{borkar2009stochastic} variants of the Bellman equation \cite{bellman1966dynamic} in MDPs. For example, the popular Q-learning algorithm \cite{watkins1992q} can be viewed as a stochastic fixed point iteration to solve the Q-Bellman equation. Therefore, we believe that in order to improve the performance of RL algorithms, a promising first step would be to propose faster algorithms for solving MDPs when the model information is known. To this end, we propose a second order value iteration technique that has global convergence, which is a desirable property.
In \cite{reetz1973solution}, the successive over-relaxation technique is applied to the Bellman equation to obtain a faster value iteration algorithm. In this work, we propose a Generalized Second Order Value Iteration (G-SOVI) method for computing the optimal value function and policy when the model information is known. This is achieved by the application of Newton-Raphson method to the Successive relaxation variant of the Q-Bellman Equation (henceforth denoted as SQBE). The key differences between G-SOVI and standard SOVI algorithms are the incorporation of relaxation parameter $w$ and the construction of single-stage reward as discussed in Section \ref{propalg}.

Note that we cannot directly apply the Newton-Raphson method to SQBE because of the presence of $\max(.)$ operator in the equation, which is not differentiable. Therefore, we approximate the $\max$ operator by a smooth function $g_N$ \cite{nesterov2005smooth}, where $N$ is a given parameter. This approximation allows us to apply the second order method thereby ensuring faster rate of convergence.

The solution obtained by our second order technique on the modified SQBE may be different from the solution of the original MDP problem because of the approximation of the $\max$ operator by $g_N$. However, we show that our proposed algorithm converges to the actual solution as $N \xrightarrow{} \infty$. 

We show through numerical experiments that given a finite number of iterations, our proposed algorithm computes a solution that is closer to the actual solution faster when compared to that obtained from standard value iteration. Moreover, under a special structure of MDP, the solution is better than standard second-order value iteration \cite{rust1994structural}. 

%% file: related-work.tex
\section{Related Work And Our Contributions}
Value Iteration and Policy Iteration are two classical numerical techniques employed for solving MDP problems. In \cite{puterman1979convergence}, it has been shown that Newton-Kantorovich\footnote{Also known as Newton-Raphson method} method applied to the exact Bellman equation gives rise to the policy iteration scheme. In (Section 2.5 of \cite{rust1994structural}), a second-order value iteration technique (we refer to it as standard SOVI) is proposed by applying Newton-Kantorovich method to the smooth (soft-max) Bellman equation and remarks about second-order rate and global convergence are provided. However, convergence analysis is not discussed for the SOVI technique. Approximate Newton methods have been proposed in \cite{furmston2016approximate} for policy optimization in MDPs. A detailed analysis of the Hessian of the objective function is provided and algorithms are derived. In recent times, smooth Bellman equation has been successfully used in the development of many RL algorithms. For instance, in \cite{haarnoja2017reinforcement}, a soft Q-learning algorithm has been proposed that learns the maximum entropy policies. The algorithm makes use of the smooth Q-Bellman equation with an additional entropy term. In \cite{dai2017sbeed}, SBEED (Smoothed Bellman Error Embedding) algorithm has been proposed which computes the optimal policy by formulating the smooth Bellman equation as a primal-dual optimization problem. 
In \cite{devraj2017zap}, a matrix-gain learning algorithm namely Zap Q-learning has been proposed which is seen to have similar performance as the Newton-Raphson method. Very recently, an accelerated value iteration technique is proposed in \cite{goyal2019first} by applying Nestorov's accelerated gradient technique to value iteration.  

We now summarize the main contributions of our paper:
\begin{itemize}
    \item We propose a generalized second order Q-value iteration algorithm that is derived from the successive relaxation technique as well as the Newton-Raphson method. In fact, we show that standard SOVI is a special case of our proposed algorithm.
    \item We prove the global convergence of our algorithm and provide a second order convergence rate result.
    \item We derive a bound on the error defined in terms of the value function obtained by our proposed method and the actual value function and show that the error vanishes asymptotically.
    \item Through experimental evaluation, we further confirm that our proposed technique provides a better near-optimal solution compared to that of the value iteration procedure when run for the same (finite) number of iterations.
\end{itemize}


%% file: background.tex
\section{Background and Preliminaries}
A discounted reward Markov Decision Process (MDP) is characterized via a tuple $(S,A,p,r,\gamma)$ where $S=\{1,2,\cdots,i,\ldots,j,\cdots,M \}$ denotes the set of states, $A = \{a_1,\ldots,a_{K}\}$ denotes the set of actions, $p$ is the transition probability rule i.e., $p(j|i,a)$ denotes the probability of transition from state $i$ to state $j$ when action $a$ is chosen. Also, $r(i,a,j)$ denotes the single-stage reward obtained in state $i$ when action $a$ is chosen and the next state is $j$. Finally, $0 \leq \g < 1$ denotes the discount factor. The objective in an MDP is to learn an optimal policy $\pi: S \xrightarrow{} A$, where $\pi(i)$ denotes the action to be taken in state $i$, that maximizes the cumulative discounted reward objective given by:
\begin{align} \label{eq-1}
    \E \Big[ \sum_{t = 0}^{\infty} \g^{t}r(s_{t},\pi(s_{t}),s_{t+1}) \mid s_{0} = i \Big].
\end{align}
In \eqref{eq-1}, $s_{t}$ is the state at time $t$ and $\E[.]$ is the expectation taken over the entire trajectory of states obtained over times $t = 1,\ldots,\infty$. Let $V^*(.)$ be the value function with $V^*(i)$ being the value of state $i$ that represents the total discounted reward obtained starting from state $i$ and following the optimal policy $\pi$. The value function can be obtained by solving the Bellman equation \cite{bertsekas1996neuro} given by:
\begin{align}\label{v-eq}
    V^*(i)=\max_{a \in A} \Big \{ \sum_{j=1}^{M} p(j|i,a) \big{(}r(i,a,j)+\g V^*(j)\big{)} \Big \} , ~ \forall i \in S.
\end{align}
We assume here for simplicity that all actions are feasible in every state.
Value iteration is a popular numerical scheme employed to obtain the value function and so the optimal policy. It works as follows: An initial estimate of the value function $V_0$ is selected arbitrarily and a sequence of $V_{n}, ~ n \geq 1$ is generated in an iterative fashion as below:
\begin{align}\label{vi-eqn}
    V_{n}(i) =& \max_{a \in A} \Big\{  \displaystyle\sum_{j=1}^{M} p(j|i,a)\big{(} r(i,a,j)+ \g V_{n-1}(j) \big{)} \Big\}, \nonumber \\ & ~n \geq 1, \forall i \in S.
\end{align}
Let $\zeta$ denote the set of all bounded functions from $S$ to $\R$. Note that equation \eqref{v-eq} can be rewritten as:
\begin{align}\label{fp-vi}
    V^* = TV^*,
\end{align}
where the operator $T: \zeta \xrightarrow{} \zeta$ is  defined by: $$(TV)(i) = \max_{a \in A} \Big\{ r(i,a)+\g \displaystyle\sum_{j=1}^{M} p(j|i,a)V(j) \Big\},$$ and 
$r(i,a)=\displaystyle\sum_{j=1}^{M} p(j|i,a)r(i,a,j) $ is the expected single-stage reward in state $i$ when action $a$ is chosen. 
It is easy to see that $T$ is a sup-norm contraction map with contraction factor $\g$, i.e., the discount factor.  Therefore, from the contraction mapping theorem, it is clear that the value iteration scheme given by equation \eqref{vi-eqn} converges to the optimal value function, i.e.,
\begin{align}
    V^* = \lim_{n \xrightarrow{} \infty} V_{n} = TV^*.
\end{align}
Let $Q^{*}(i,a)$ with $(i,a) \in S \times A$, be defined as:
\begin{align}\label{ql-eq}
    Q^{*}(i,a) := r(i,a)+\g \sum_{j=1}^{M} p(j|i,a)V^*(j).
\end{align}
Here $Q^{*}(i,a)$ is the optimal Q-value function associated with state $i$ and action $a$. It denotes the total discounted reward obtained starting from state $i$ upon taking action $a$ and following the optimal policy in subsequent states. Then from \eqref{v-eq}, it is clear that
\begin{align}
    V^*(i) = \max_{a \in A}Q^{*}(i,a).
\end{align}
Therefore, the equation \eqref{ql-eq} can be re-written as follows:
\begin{align}\label{ql-eq2}
    Q^{*}(i,a) = r(i,a) + \g \sum_{j=1}^{M} p(j|i,a) \max_{b \in A}Q^{*}(j,b).
\end{align}
This is known as the Q-Bellman equation. 
We obtain the optimal policy by letting
\begin{align}
    \pi(i) = \arg \max_{a \in A}Q^{*}(i,a).
\end{align}
In \cite{reetz1973solution}, a modified value iteration algorithm is proposed based on the idea of successive relaxation. Let us define 
\begin{align}\label{wstar-def}
    w^* = \frac{1}{1-\gamma\displaystyle\min_{i,a}p(i|i,a)}. 
\end{align}
Note that $w^* \geq 1$.
For $0 < w\leq w^*,$ we define a modified Bellman operator as follows:
\begin{align}\label{soT}
    T_w(V) = w(TV) + (1-w)V,
\end{align}
where $w$ is called the `relaxation' parameter. It is easy to see that the fixed point of $T_w$ is also the optimal value function of the MDP (fixed point of $T$). Moreover, it is shown in \cite{reetz1973solution} that the contraction factor of $T_w$ is $1-\gamma+\gamma w$. Under a special structure of the MDP, i.e., with $p(i| i,a) > 0, ~ \forall i,a$, we have $w^* > 1$ (strictly greater than $1$). Then, the relaxation parameter $w$ can be chosen in three possible ways:
\begin{enumerate}
    \item If $0<w<1,$  then the contractor factor of $T_w$ is more than the contraction factor of $T$.
    \item If $w=1,$ then $T = T_w$ and hence the contraction factors of both the operators are same.
    \item If $1 < w \leq w^*,$ the contraction factor of $T_w$ is less than the contraction factor of $T$. This implies that the fixed point iteration utilizing \eqref{soT} generates the optimal value function faster than the standard value iteration.
\end{enumerate}
In \cite{kamanchi2019successive}, a successive relaxation Q-Bellman equation (we call the Generalized Q-Bellman equation) is constructed as follows:
\begin{align}\label{QsoT}
    Q_w(i,a) = w(r(i,a) +& \g \sum_{j=1}^{M} p(j|i,a) \max_{b \in A}Q_w(j,b)) \nonumber \\  &+ (1-w)\max_{c \in A}Q_w(i,c),
\end{align}
where $0 < w \leq w^*$. It has been shown that, although the Q-values obtained by \eqref{QsoT} can be different from the optimal Q-values, the optimal value functions are still the same. That is, 
\begin{align}
    \max_{a \in A}Q_w(i,a) = \max_{a \in A}Q^{*}(i,a), ~ \forall i \in S.
\end{align}

The Generalized Q-values ($Q_w$ in \eqref{QsoT}) are obtained as follows. An initial estimate $Q_0$ of $Q_w$ is arbitrarily selected and a sequence of $Q_n, ~ n \geq 1$ is obtained according to:

\begin{align}\label{IterQsoT}
Q_{n}(i,a) =&  w(r(i,a) + \g \sum_{j=1}^{M} p(j|i,a) \max_{b \in A}Q_{n-1}(j,b)) \nonumber \\  &+ (1-w)\max_{c \in A}Q_{n-1}(i,c), ~ \forall (i,a).
\end{align}

It is shown in \cite{kamanchi2019successive} that, the Q-values obtained by \eqref{IterQsoT} converge to the generalized Q-values $Q_w$. 
In this way, we obtain optimal value function and optimal policy using the successive relaxation Q-value iteration scheme. In this work, our objective is to approximate the generalized Q-Bellman equation and apply the Newton-Raphson second order technique to solve for the optimal value function. Recall that we cannot apply the second order method directly to the equation \eqref{QsoT} as the $\max(.)$ operator on the RHS is not differentiable. Before we propose our algorithm, we briefly discuss the Newton's second order technique  \cite{ortega1970iterative} for solving a non-linear system of equations.

Consider a function $F:\R^{d} \xrightarrow{} \R^{d}$ that is twice differentiable.  Suppose we are interested in finding a root of $F$ i.e., a point $x$ such that $F(x) = 0$. The Newton-Raphson method can be applied to find a solution here. We select an initial point $x_0$ and then proceed as follows:
\begin{align}\label{nw-method}
    x_n = x_{n-1} - J_F^{-1}(x_{n-1})F(x_{n-1}), ~ n \geq 1,
\end{align}
where $J_F(x)$ is the Jacobian of the function $F$ evaluated at point $x$. Under suitable hypotheses it can be shown that the procedure \eqref{nw-method} leads to second order convergence.

In the next section, we construct a function $F$ for our problem and apply the Newton-Raphson method to find the optimal value function and policy pair. 

%% file: proposed-algorithm.tex
\section{Proposed Algorithm}\label{propalg}
We construct our modified SQBE as follows. We first approximate the $\max(.)$ operator, i.e., the function $f(x)=\max^{d}_{i=1}{x_{i}}$, where $x = (x_1,\ldots,x_d)$, with 
$g_{N}(x)=\frac{1}{N}\log\displaystyle\sum^{d}_{i=1}e^{Nx_{i}},$ as the $\max(.)$ operator is not differentiable. We note here that $g_{N}(x)$ is a smooth approximation of $\max$ operator $f(x)$ as shown in the Lemma \ref{l3}. Then the equation \eqref{IterQsoT} can be rewritten as follows:
\begin{align}\label{mqvi-eqn}
    Q_{n}(i,a) = w\bigg{(} & r(i,a)+\g \displaystyle\sum_{j=1}^{M} p(j|i,a)\frac{1}{N}\log\displaystyle\sum^{|A|}_{b=1}e^{NQ_{n-1}(j,b)}\bigg{)}\nonumber \\
    &+(1-w)\frac{1}{N}\log\displaystyle\sum^{|A|}_{b=1}e^{NQ_{n-1}(i,b)}, ~n \geq 1,
\end{align}
starting with an initial $Q_0$ (arbitrarily chosen in general).
Therefore our modified Successive Q-Bellman (SQB) operator $U:\R^{|S| \times |A|} \xrightarrow{} \R^{|S| \times |A|} $ is defined as follows. For $0 < w \leq w^*$,
\begin{align}\label{mbo}
    UQ(i,a)= w\bigg{(} & r(i,a)+\g \displaystyle\sum_{j=1}^{M} p(j|i,a)\frac{1}{N}\log\displaystyle\sum^{|A|}_{b=1}e^{NQ(j,b)}\bigg{)}\nonumber \\
    &+(1-w)\frac{1}{N}\log\displaystyle\sum^{|A|}_{b=1}e^{NQ(i,b)}.
\end{align}
The numerical scheme \eqref{mqvi-eqn} is thus
\begin{align*}
    Q_n(i,a) = UQ_{n-1}(i,a).
\end{align*}
Finally, by an application of the Newton-Raphson method to $U$, our Generalized Second Order Value Iteration (G-SOVI) is obtained as described in Algorithm \ref{alg:SOVI}. Note that setting $w=1$ in Step 4 of the algorithm yields the standard SOVI algorithm.  
\begin{algorithm}[ht]
\caption{Generalized Second Order Value Iteration (G-SOVI) }\label{alg:SOVI}
\hspace*{\algorithmicindent} \textbf{Input:}\\ 
\hspace*{\algorithmicindent} $(S,A,p,r,\g)$: MDP model \\
\hspace*{\algorithmicindent}
$0 < w \leq w^*$: Prescribed relaxation parameter \\
\hspace*{\algorithmicindent} $Q_0$: Initial $Q$-vector \\
\hspace*{\algorithmicindent} $N$: Prescribed approximation factor\\
\hspace*{\algorithmicindent} Iter: Total number of iterations\\
\hspace*{\algorithmicindent} \textbf{Output:}  $Q_{\text{Iter}}$
\begin{algorithmic}[1]
\Procedure{G-SOVI:}{}
\While{$n < $ Iter}
\State compute $|S\times A| \times |S \times A|$ matrix $J_{U}(Q_n)$. The $((i,a),(k,c))^{\text{th}}$ entry is given by \\
$J_{U}(Q_{n})((i,a),(k,c))=$\[ \begin{cases} 
      w\g p(k|i,a) \frac{e^{NQ_n(k,c)}}{\displaystyle\sum_{b \in A}e^{NQ_n(k,b)}} & (k,c)\neq(i,a)\\
      (1-w+w\g p(k|i,a))\frac{e^{NQ_n(k,c)}}{\displaystyle\sum_{b \in A}e^{NQ_n(k,b)}} & (k,c)=(i,a)\\
   \end{cases}
\]
\State $Q_{n+1}=Q_n- \big{(}I-J_{U}(Q_n)\big{)}^{-1}(Q_n-UQ_{n})$
\EndWhile
\State \textbf{return} $Q_{\text{Iter}}$ 
\EndProcedure
\end{algorithmic}
\end{algorithm}
\begin{remark}
Note that in our case, the function $F$ in equation \eqref{nw-method} corresponds to $F(Q)=Q-UQ$ and $J_{F}(Q)=I-J_{U}(Q)$ is a $|S\times A| \times |S \times A|$ dimensional matrix.
\end{remark}
\begin{remark}
Note that directly computing $\big{(}I-J_{U}(Q)\big{)}^{-1}(Q-UQ)$ would involve $O(|S|^3|A|^3)$ complexity. This computation could be carried out by solving the system $(I-J_{U}(Q))Y=Q-UQ$ for $Y$ to avoid numerical stability issues. Moreover the per-iteration time complexity of the Algorithm \ref{alg:SOVI} is also $O(|S|^3|A|^3)$.
\end{remark}
\begin{remark}
Note that G-SOVI reduces to standard SOVI in the case $w=1.$ Moreover the computational complexity for both the algorithms is the same.
\end{remark}
\begin{remark}
The space required for storing the Jacobian matrix $J_{U}$ is $|S|^2|A|^2$. Hence the space complexity of Algorithm\ref{alg:SOVI} is $O(|S|^2|A|^2)$.
\end{remark}

%% file: convergence-analysis.tex
\section{Convergence Analysis}\label{conv-sec}
In this section we study the convergence analysis of our algorithm. Note that the norm considered in the following analysis is the max-norm, i.e., $\|x\|:=\max_{1\leq i \leq d}|x_{i}|.$ Throughout this section, it is assumed that the relaxation parameter $w$ satisfies $0<w\leq w^*,$ where $w^*$ is as defined in \eqref{wstar-def}.
\vspace*{0.4cm}
\begin{lemma}
Suppose $f: \R^{d} \rightarrow \R$ and $f(x):=\displaystyle\max\{x_{1},x_{2},\cdots,x_{d}\}. $ Let $g_N: \R^{d} \rightarrow \R$ be defined as 
$g_N(x):=\frac{1}{N}\log\displaystyle\sum_{i=1}^{d}e^{Nx_{i}}. $ Then $\displaystyle\sup_{x \in \R^{d}}\big{|}f(x)-g_{N}(x)\big{|} \longrightarrow 0$ as $N \longrightarrow \infty.$
\end{lemma}
\proof
Let $x_{i_{*}}=\max\{x_{1},x_{2},\cdots,x_{d}\}$ (where $i_*$ denotes the corresponding $\arg \max$). Now
\begin{align*}
     \big{|}f(x)-g_{N}(x)\big{|}
    & =\bigg{|}\max\{x_{1},x_{2},\cdots,x_{d}\}-\frac{1}{N}\log\displaystyle\sum_{i=1}^{d}e^{Nx_{i}}\bigg{|}\\
    & = \bigg{|}x_{i_{*}}-\frac{1}{N}\log\Big{[}\Big{(}\displaystyle\sum_{i=1}^{d}e^{N(x_{i}-x_{i_{*}})}\Big{)}e^{Nx_{i_{*}}}\Big{]}\bigg{|}\\
    & = \bigg{|}\frac{1}{N}\log\bigg{(}\displaystyle\sum_{i=1}^{d}e^{N(x_{i}-x_{i_{*}})}\bigg{)}\bigg{|}\\
    & \leq \bigg{|}\frac{\log d}{N}\bigg{|} \rightarrow 0 \text{ as } N \rightarrow \infty.
\end{align*}
Note that the inequality follows from the definition of $x_{i_{*}}=\max\{x_{1},x_{2},\cdots,x_{d}\} $ and the fact that $e^{N(x_{i}-x_{i_{*}})} \leq 1 \text{ for } 1 \leq i \leq d$ (since $x_i \leq x_{i^*}$ $\forall i$).
Hence $\displaystyle\sup_{x \in \R^{d}}\big{|}f(x)-g_{N}(x)\big{|} \rightarrow 0$ as $N \rightarrow \infty $ with the rate $\frac{1}{N}.$
\vspace*{0.4cm}
\begin{lemma}
\label{contraction-l2}
Let $U:\R^{|S| \times |A|} \rightarrow \R^{|S| \times |A|}$ be defined as follows.
\begin{align*}
(UQ)(i,a)= w\bigg{(} & r(i,a)+\g \displaystyle\sum_{j=1}^{M} \frac{p(j|i,a)}{N}\log\displaystyle\sum^{|A|}_{b=1}e^{NQ(j,b)}\bigg{)}\nonumber \\
    &+\frac{(1-w)}{N}\log\displaystyle\sum^{|A|}_{b=1}e^{NQ(i,b)}.
\end{align*}
Then $U$ is a max-norm contraction.
\end{lemma}
\proof
Given $(i,a)$, let $q(.|i,a):\{1,2,\cdots,|S|\}\rightarrow [0,1]$ be defined as follows. \[ q(k|i,a)= \begin{cases} 
      \frac{w\g p(k|i,a)}{(1-w+w\g)}, ~ k\neq i,\\
      \frac{1-w+w\g p(i|i,a)}{(1-w+w\g)}, ~  k=i.
   \end{cases}\]
Observe that $q$ is a probability mass function. Let $\E[.]$ denote the expectation with respect to $q$, and $\xi(j,.)$ denotes the point that lies on the line joining $P(j,.)$ and $Q(j,.)$
Now for $P,Q \in \R^{|S| \times |A|}$, we have
\begin{align*}
& \big{|}(UP)(i,a)-(UQ)(i,a) \big{|}\\
& = (1-w+w\g)\Bigg{|}\E\Bigg{[}\frac{\log\displaystyle\sum_{b=1}^{|A|}e^{NP(j,b)}}{N}-\frac{\log\displaystyle\sum_{b=1}^{|A|}e^{NQ(j,b)}}{N}\Bigg{]}\Bigg{|}\\
& = (1-w+w\g)\Bigg{|}\E\Bigg{[}\Bigg{(}\frac{e^{N\xi(j,.)}}{\displaystyle\sum_{b \in A}e^{N\xi(j,b)}}\Bigg{)}^T \Big{(}P(j,.)-Q(j,.)\Big{)}\Bigg{]}\Bigg{|}\\
&\leq(1-w+w\g)\E\Bigg{[}\Bigg{|}\Bigg{(}\frac{e^{N\xi(j,.)}}{\displaystyle\sum_{b \in A}e^{N\xi(j,b)}}\Bigg{)}^T \Big{(}P(j,.)-Q(j,.)\Big{)}\Bigg{|}\Bigg{]}\\
& \leq (1-w+w\g)\E\big{[}\max_{b}|P(j,b)-Q(j,b)| \big{]} \\
& \leq (1-w+w\g) \max_{(i,a)}|P(i,a)-Q(i,a)|\\
& =(1-w+w\g) \|P-Q\|\\
& \text{So    }  \|UP-UQ\| = \max_{(i,a)}\big{|}(UP)(i,a)-(UQ)(i,a) \big{|} \\
& \leq (1-w+w\g) \|P-Q\|.
\end{align*}
Hence $U$ is a contraction with contraction factor $(1-w+w\g)$. Here, the second equality follows from an application of mean value theorem in multivariate calculus. 
\vspace*{0.4cm}

\begin{lemma} \label{l3}
Let $Q_w$ be as in equation \eqref{QsoT} and $Q'$ be fixed point of $U$ respectively.
Then $\|Q_w-Q'\|\leq \frac{(1-w+w\g)}{Nw(1-\g)} \log(|A|).$
\end{lemma}
\proof
From equation \eqref{QsoT}, we have
$$Q_w(s,a)=wr(s,a)+(1-w+w\g) \E\bigg{[}\displaystyle\max_{b \in A} Q_w(Z,b)\bigg{]}.$$
Now $Q'$ is the unique fixed point of $U$ (unique by virtue of Lemma \ref{contraction-l2}), so
$$Q'(s,a)=wr(s,a)+ (1-w+w\g) \E\bigg{[}\frac{1}{N}\log\displaystyle\sum_{b \in A}
e^{NQ'(Z,b)}\bigg{]},$$
where $Z$ is a random variable with probability mass function as $q$ and the expectation above is taken with respect to the law given by probability mass function $q$. Let $c= \arg\max_{b \in A}Q'(Z,b)$ i.e. $Q'(Z,c)=\max_{b \in A}Q'(Z,b).$ 
Now
\begin{align*}
    & \big{|}Q_w(s,a)-Q'(s,a)\big{|} \\
   = & \Bigg{|}(1-w+w\g)\E\bigg{[}\max_{b \in A} Q_w(Z,b)-\frac{1}{N}\log\displaystyle\sum_{b \in A}
e^{NQ'(Z,b)}\bigg{]}\Bigg{|}\\
 = & (1-w+w\g)\Bigg{|} \E\bigg{[}\max_{b \in A}Q_w(Z,b)\\
 &\hspace{0.25cm} -\max_{b \in A}Q'(Z,b)-\frac{1}{N}\log\Big{(}\displaystyle\sum_{b \in A}e^{N\big{(}Q'(Z,b)-Q'(Z,c)\big{)}}\Big{)}
\bigg{]} \Bigg{|}\\
\leq & (1-w+w\g) \E\Bigg{[}\bigg{|}\max_{b \in A}Q_w(Z,b) \\
&\hspace{0.25cm} -\max_{b \in A}Q'(Z,b)-\frac{1}{N}\log\Big{(}\displaystyle\sum_{b \in A}e^{N\big{(}Q'(Z,b)-Q'(Z,c)\big{)}}\Big{)}
\bigg{|}\Bigg{]}
\end{align*}
\begin{align*}
\leq & (1-w+w\g) \E\Bigg{[}\bigg{|}\max_{b \in A}Q_w(Z,b)-\max_{b \in A}Q'(Z,b)\bigg{|}\\
&\hspace{2cm}+\bigg{|}\frac{1}{N}\log\Big{(}\displaystyle\sum_{b \in A}e^{N\big{(}Q'(Z,b)-Q'(Z,c)\big{)}}\Big{)}
\bigg{|}\Bigg{]} \\
\leq & (1-w+w\g) \|Q_w-Q'\|+\frac{(1-w+w\g)}{N}\log|A|.
\end{align*}
\begin{align*}
& \text{Hence }\\
&\|Q_w-Q'\| \\ 
&\hspace{1cm} \leq (1-w+w\g) \|Q_w-Q'\|+\frac{(1-w+w\g)}{N}\log|A| \\
& \implies \|Q_w-Q'\| \leq \frac{(1-w+w\g)}{Nw(1-\g)} \log|A|.
\end{align*}
This completes the proof. This lemma shows that the approximation error $\|Q_w-Q'\| \rightarrow 0$ as $N \rightarrow \infty.$

\vspace*{0.4cm}
\begin{remark}
It is easy to see that in the case of $w>1$
\begin{align*}
    \frac{(1-w+w\g)}{w}<\g.
\end{align*}
This shows that the approximation error of G-SOVI is smaller than standard SOVI in the case of $w>1$.
\end{remark}

We now invoke the following theorem from \cite{ortega1970iterative} to show the global convergence of our second order value iteration.
\vspace*{0.4cm}
\begin{theorem}[Global Newton Theorem]
\label{GNT}
Suppose that $F: \R^{d}\rightarrow \R^{d}$ is continuous, component-wise concave on $\R^{d}$, differentiable and that $F'(x)$ is non-singular and $F'(x)^{-1} \geq 0$, i.e. each entry of $F'(x)^{-1}$ is non-negative, for all $x \in \R^{d}.$ Assume, further, that $F(x)=0$ has a unique solution $x^{*}$ and that $F'$ is continuous on $\R^{d}.$ Then for any $x_{0} \in \R^{d}$ the Newton iterates given by \eqref{nw-method} converge to $x^{*}.$
\end{theorem}
\vspace*{0.4cm}
\begin{remark}
\label{r3}
Note that the above theorem is stated for convex $F$ in \cite{ortega1970iterative}. However, the theorem holds true even for concave $F$. 
\end{remark}
\vspace*{0.4cm}
\begin{theorem}
Let $Q'$ be the fixed point of the operator $U$.
G-SOVI converges to $Q'$ for any choice of initial point $Q_{0}.$
\end{theorem}
\proof
G-SOVI computes the zeros of the equation $Q-UQ=0.$
So we appeal to Theorem \ref{GNT} with the choice of $F$ as $I-U:\R^{|S|\times|A|}\rightarrow \R^{|S|\times|A|}$ where $(I-U)(Q)(i,a)=Q(i,a)-w r(i,a)-(1-w+\g w) \E\bigg{[}\frac{1}{N}\log\displaystyle\sum_{b \in A}
e^{NQ(Z,b)}\bigg{]}.$
It is enough to verify the hypothesis of Theorem \ref{GNT} for $I-U.$ Clearly $I-U$ is continuous, component-wise concave and differentiable with $(I-U)'(Q)= I-J_{U}(Q)$ where 
\begin{align*}
J_{U}(Q)((i,a),(k,c))=(1-w+w\g) q(k|i,a) \frac{e^{NQ(k,c)}}{\displaystyle\sum_{b \in A}e^{NQ(k,b)}}.
\end{align*}
Note that $(I-U)$ is a $|S\times A| \times |S \times A|$ dimensional matrix with $1\leq i,k \leq |S|$ and $1 \leq a,c \leq |A|$. Now observe that
\begin{itemize}
    \item each entry in the $(i,a)^{\text{th}}$ row is non-negative.
    \item the sum of the entries in the $(i,a)^{\text{th}}$ row is
\begin{align*}
 &\displaystyle\sum^{|S|}_{k=1}\displaystyle\sum^{|A|}_{c=1} (1-w+w\g) q(k|i,a) \frac{e^{NQ_n(k,c)}}{\displaystyle\sum_{b \in A}e^{NQ_n(k,b)}}\\
 &=(1-w+w\g).    
\end{align*}
\end{itemize}
So $J_{U}(Q)=(1-w+w\g)\Phi$ for a $|S\times A| \times |S \times A|$ dimensional transition probability matrix $\Phi$. It is easy to see that $(I-J_{U}(Q))^{-1}$ exists (see Lemma \ref{l4}) with the power series expansion
\begin{align*}
    \big{(}I-J_{U}(Q)\big{)}^{-1}= \displaystyle\sum^{\infty}_{r=0} (1-w+w\g)^{r} \Phi^{r}.
\end{align*}
Moreover, since each entry in $\Phi$ is non-negative, $\Phi \geq0$. Hence $\big{(}I-J_{U}(Q)\big{)}^{-1}\geq 0.$ Also from lemma \ref{contraction-l2} it is clear  that the equation $Q-UQ=0$ has a unique solution. This completes the proof.
\vspace*{0.4cm}
\begin{lemma}\label{l4}
$\left\|\big{(}I-J_{U}(Q)\big{)}^{-1}\right\| \leq \frac{1}{w(1-\g)}$
\end{lemma}
\label{r4}
\proof
Note that 
\begin{align*}
I-J_{U}(Q)=I-(1-w+w\g)\Phi, 
\end{align*} 
for a given transition probability matrix $\Phi$. Now suppose that $\lambda$ is an eigen-value of $I-(1-w+w\g)\Phi$ then
\begin{align*}1-(1-w+w\g)<\lambda\leq 1+(1-w+w\g).\end{align*} From $1-(1-w+w\g)>0$, we have
\begin{align*}
    0 \notin \sigma(I-J_{U}(Q)),
\end{align*}
where $\sigma(I-J_{U}(Q))$ is the spectrum of $I-J_{U}(Q)$. Hence for any $Q$, $\big{(}I-J_{U}(Q)\big{)}^{-1}$ exists and we have
\begin{align*}
    \left\|\big{(}I-J_{U}(Q)\big{)}^{-1}\right\| & \leq \frac{1}{1-(1-w+w\g)} \\ & =\frac{1}{w(1-\g)}.
\end{align*}
This completes the proof.\\ 
The following theorem is an adaptation from \cite{ortega1970iterative}.
\begin{theorem}
G-SOVI has second order convergence.
\end{theorem}
\proof
Recall that $F(Q)=Q-UQ.$ Let $Q^*$ be the unique solution of $F(Q)=0$ and $\{Q_{n}\}$ be the sequence of iterates generated by G-SOVI.
Define $e_{n}=\|Q_{n}-Q^*\|$ and $G(Q)=Q-F'(Q)^{-1}F(Q)$. As $Q^*$ satisfies $Q^* = UQ^*$, it is a fixed point of $G$. It is enough to show that $e_{n+1}\leq ke^2_{n}$ for a constant $k.$
It can be shown that for our particular choice of $F$, $F'$ is Lipschitz (with Lipschitz constant, say, $L$). 
\begin{align*}
    & \|F'(Q)-F'(Q^*)\| \leq  L \|Q-Q^*\| \\
    & \implies \|F(Q)-F(Q^*)-F'(Q^*)(Q-Q^*)\|\leq \frac{L}{2} \|Q-Q^*\|^2 \\ 
    & \text{(by an application of the fundamental theorem of calculus).}
\end{align*}
Utilizing the above properties we have
\begin{align*}
    e_{n+1} = & \|Q_{n+1}-Q^*\| \\
    = & \|G(Q_{n})-G(Q^*)\| \\
    = & \|Q_{n}-F'(Q_{n})^{-1}F(Q_{n})- Q^*\|\\
    \leq & \big{\|}F'(Q_{n})^{-1}\big{[}F(Q_{n})-F(Q^*)-F'(Q^*)(Q_{n}-Q^*)\big{]}\big{\|} \\ &+\big{\|}F'(Q_{n})^{-1}\big{[}F'(Q_{n})-F'(Q^*)\big{]}(Q_{n}-Q^*)\big{\|} \\
    \leq & \|F'(Q_{n})^{-1}\|\big{\|}\big{[}F(Q_{n})-F(Q^*)-F'(Q^*)(Q_{n}-Q^*)\big{]}\big{\|} \\ &+\|F'(Q_{n})^{-1}\|\big{\|}\big{[}F'(Q_{n})-F'(Q^*)\big{]}(Q_{n}-Q^*)\big{\|} \\   
    \leq & \frac{3}{2}\beta \|Q_{n}-Q^*\|^2 \\
    = & k e^2_{n},
\end{align*}
where $\beta=L\|F'(Q)\|^{-1} \leq \frac{L}{w(1-\g)}$ (from Lemma \ref{l4}) and $k=\frac{3}{2}\beta$.

%% file: experiments.tex
\section{Experiments}
\begin{figure}
    \centering
    \includegraphics[scale = 0.5]{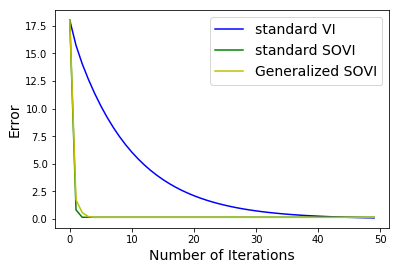}
    \caption{Error vs Number of iterations on setting with $100$ states and $10$ actions with $w = w^*$ for Generalized SOVI (G-SOVI).}
    \label{fig2}
\end{figure}
\begin{table}[h!]
\renewcommand{\arraystretch}{1.3}
\centering
\begin{tabular}{|l|c|c|c|}
\hline
\textbf{Value of N}              & \textbf{Standard Value}    & \textbf{Standard } & \textbf{G-SOVI} \\ 
          & \textbf{Iteration}    &  \textbf{SOVI} &  \\ \hline

\textbf{N=20} & \multirow{4}{*}{0.1009 $\pm$ 0.0026} & 0.1205 $\pm$ 0.0372    & 0.1093 $\pm$ 0.0818       \\ \cline{1-1} \cline{3-4} 
\textbf{N=25} &                                      & 0.0822 $\pm$ 0.0273    & 0.0648 $\pm$ 0.0217       \\ \cline{1-1} \cline{3-4} 
\textbf{N=30} &                                      & 0.0611 $\pm$ 0.0211    & 0.0494 $\pm$ 0.017        \\ \cline{1-1} \cline{3-4} 
\textbf{N=35} &                                      & 0.0484 $\pm$ 0.0168    & 0.0397 $\pm$ 0.0136       \\ \hline
\end{tabular}
\caption{Comparison of Average Error for different values of $N$ on $10$ states and $5$ actions setting at the end of $50$ iterations. For the G-SOVI algorithm, the relaxation parameter is chosen to be the optimal relaxation parameter $w^*$, i.e., $w = w^*$. }
\label{table1}
\end{table}
\begin{table}[h!]
\renewcommand{\arraystretch}{1.3}
\centering
\begin{tabular}{|c|c|}
\hline
\textbf{Value of $w$}                                             & \textbf{G-SOVI}     \\ \hline
\begin{tabular}[c]{@{}c@{}}$w = 1$\\ (Standard SOVI)\end{tabular} & 0.04838 $\pm$ 0.017 \\ \hline
$w = 1.00001$                                                     & 0.04838 $\pm$ 0.017 \\ \hline
$w = 1.0001$                                                      & 0.04837 $\pm$ 0.017  \\ \hline
$w = 1.001$                                                       & 0.04830 $\pm$ 0.017 \\ \hline
$w = 1.01$                                                        & 0.0476 $\pm$ 0.017  \\ \hline
$w = 1.05$                                                        & 0.0448 $\pm$ 0.016  \\ \hline
$w = 1.1$                                                         & 0.0417 $\pm$ 0.014  \\ \hline
$w = w^*$                                                         & 0.0397 $\pm$ 0.014 \\ \hline
\end{tabular}
\caption{Comparison of Average Error in G-SOVI for different values of $w$ on $10$ states and $5$ actions setting at the end of $50$ iterations. The value of $N$ is 35.}
\label{table-new}
\end{table}
\begin{table*}[h!]
\renewcommand{\arraystretch}{1.3}
\centering
\begin{tabular}{|c|c|c|c|}
\hline
\textbf{Setting}                    & \textbf{\begin{tabular}[c]{@{}c@{}}Standard \\ Value Iteration\end{tabular}} & \textbf{Standard SOVI} & \textbf{G-SOVI} \\ \hline
\textbf{States = 30, Actions= 10}   & 6.471 $\pm$ 0.07                                                                        & 0.087 $\pm$ 0.01                    & 0.079 $\pm$ 0.01                       \\ \hline
\textbf{States = 50, Actions = 10}  & 6.587 $\pm$ 0.07                                                                       & 0.114 $\pm$ 0.01                  & 0.108 $\pm$ 0.01                    \\ \hline
\textbf{States = 80, Actions = 10}  & 6.754 $\pm$ 0.03                                                                      & 0.141 $\pm$ 0.01                   & 0.136 $\pm$ 0.01                     \\ \hline
\textbf{States = 100, Actions = 10} & 6.772 $\pm$ 0.03                                                                         & 0.152 $\pm$ 0.01                   & 0.148 $\pm$ 0.01                      \\ \hline
\end{tabular}
\caption{Comparison of Average Error across four settings at the end of $10$ iterations with $N=35$. For the G-SOVI algorithm, the relaxation parameter is chosen to be the optimal relaxation parameter $w^*$, i.e., $w = w^*$. }
\label{table2}
\end{table*}
\begin{table*}[h!]
\renewcommand{\arraystretch}{1.3}
\centering
\begin{tabular}{|c|c|c|c|}
\hline
\textbf{Setting}                    & \textbf{\begin{tabular}[c]{@{}c@{}}Standard \\ Value Iteration\end{tabular}} & \textbf{Standard SOVI} & \textbf{G-SOVI} \\ \hline
\textbf{States = 30, Actions= 10}   & 0.0008 $\pm$ 0.00                                                                       & 0.0154 $\pm$ 0.01                   & 0.0267 $\pm$ 0.01                      \\ \hline
\textbf{States = 50, Actions = 10}  &  0.0009 $\pm$ 0.00                                                                        & 0.0242 $\pm$ 0.00                  & 0.0488 $\pm$ 0.00                    \\ \hline
\textbf{States = 80, Actions = 10}  & 0.0011 $\pm$ 0.00                                                                       &  0.0532 $\pm$ 0.00                  &  0.0988 $\pm$ 0.01                     \\ \hline
\textbf{States = 100, Actions = 10} &  0.0026 $\pm$ 0.00                                                                         &  0.1202 $\pm$ 0.01                  &  0.1343 $\pm$ 0.01                     \\ \hline
\end{tabular}
\caption{Per-iteration Execution time of algorithms across four settings in seconds, with the relaxation parameter in G-SOVI chosen as $w=w^*$.}
\label{table3}
\end{table*}
\begin{table*}[h!]
\centering
\renewcommand{\arraystretch}{1.3}
\begin{tabular}{|c|c|c|c|c|}
\hline
\textbf{Configuration} & \textbf{\begin{tabular}[c]{@{}c@{}}Computational Time \\ (in seconds)\end{tabular}} & \textbf{\begin{tabular}[c]{@{}c@{}}Standard \\ Value Iteration\end{tabular}} & \textbf{Standard SOVI} & \textbf{G-SOVI} \\ \hline
10 States, 5 Actions   & 0.01                                                                                & 25.485 $\pm$ 2.21                                                            & 3.930$\pm$ 0.92        & 3.885 $\pm$ 0.94          \\ \hline
20 States, 5 Actions   & 0.02                                                                                & 18.291 $\pm$ 0.77                                                            & 5.444 $\pm$ 0.51       & 5.473 $\pm$ 0.50          \\ \hline
30 States, 5 Actions   & 0.03                                                                                & 7.327 $\pm$ 0.20                                                             & 7.111 $\pm$ 0.32       & 7.118 $\pm$ 0.33          \\ \hline
\end{tabular}
\caption{Average Error vs Computational Time (rounded off to the nearest millisecond). Initial Q-values for algorithms are assigned random integers between $60$ and $70$. The discount factor is set to $0.99$. G-SOVI is run with $w = 1.00001$.}
\label{table4}
\end{table*}
In this section, we describe the experimental results of our proposed G-SOVI algorithm and compare the same with standard SOVI and value iteration. For this purpose, we use python MDP toolbox \cite{pymdp} for generating the MDP and implementing standard value iteration \footnote{The code for our experiments is available at: \url{https://github.com/raghudiddigi/G-SOVI}}. The generated MDPs satisfy $p(i|i,a) > 0, ~ \forall i,a$ in order to ensure that $w^* >1$. We consider the error as defined below to be the metric for comparison between algorithms. Error for a given algorithm at iteration $i$, denoted $E(i)$, is calculated as follows. We collect the max-norm difference between the optimal value function and the value function estimate at iteration $i$. That is, 
\begin{align*}
    E(i) = \|V^{*} - Q^{i}(.,a^{*})\|_{\infty},
\end{align*}
where $V^*$ is the optimal value function of the MDP and $Q^i(.,.)$ is the Q-value function estimate of MDP at iteration $i$. Also, for any state $j$, $a^*_j = \displaystyle \arg \max_{a \in A} Q^{i}(j,a)$.


First, we generate $100$ independent MDPs each with $10$ states, $5$ actions and we set the discount factor to be $0.9$ in each case. We run all the algorithms for $50$ iterations. The initial Q-values of the algorithms are assigned random integers between 10 and 20 (which are far away from the optimal value function). In Table \ref{table1}, we indicate the average error value (error averaged over $100$ MDPs) at the end of $50$ iterations for all the algorithms, wherein for G-SOVI, we set $w=w^*$ as the relaxation parameter. We observe that standard SOVI and G-SOVI with $N=25,30,$ and $35$ have low error at the end of $50$ iterations compared to the standard value iteration. Moreover, the average error is the least for our proposed G-SOVI algorithm. 
Also, we find that higher the value of $N$, the smaller is the error between the G-SOVI value function and the optimal value function.

In Table \ref{table-new}, we report the performance of G-SOVI for different values of feasible successive relaxation parameters $w$ across the same $100$ MDPs generated previously (in Table \ref{table1}). The optimal successive relaxation parameter $w^*$ here lies between $1.1$ and $1.5$. Recall that G-SOVI exhibits faster convergence for any value of $w$ that satisfies $1 < w \leq w^*$ when compared to standard SOVI (first row of Table \ref{table-new}). From Table \ref{table-new}, we can conclude that G-SOVI with $w \in (1,w^*]$ performs at least as fast as the standard SOVI. Moreover, the higher the value of $w$, the better is the performance, when the algorithm is run for a sufficient number of iterations.

In Table \ref{table2}, we present the results of the three algorithms on four different settings, averaged over $10$ MDPs. The standard SOVI and G-SOVI are run with $N=35$. All the algorithms are run for $10$ iterations. We observe that standard SOVI and G-SOVI have low error compared to the standard value iteration. Moreover, the difference here is much more pronounced than in Table \ref{table1}, where algorithms are run for $50$ iterations. Recall that the SOVI and G-SOVI algorithms with a fixed $N$ give near-optimal value functions. The advantage of using our proposed algorithms is that the Q-value iterates converge to the near-optimal Q-values rapidly. This can also be observed in Figure \ref{fig2}, where we present the convergence of algorithms over $50$ iterations on $100$ states and $10$ actions setting. The SOVI and G-SOVI algorithms converge rapidly to a value and stay constant. In fact, we observe here that the error is less than that obtained by the standard value iteration till $45$ iterations. Moreover, G-SOVI computes a solution that gives lower error as compared to that obtained by SOVI.

In Table \ref{table3}, we indicate the per-iteration execution time of our algorithms across the four settings considered above. We can see that, due to Hessian inversion operation in the second-order techniques, standard and  G-SOVI algorithms take more time compared to the standard value iteration algorithm.

Recall that the advantage of second-order methods is that even though the per-iteration computation is higher compared to the first-order methods, the total number of iterations needed to achieve a desired error threshold is much lower in general. Hence, they are capable of achieving lower error in the same computational time. We demonstrate this in Table \ref{table4} for three settings. We select the parameters of this experiment (i.e., number of states and actions, values of $N$, $w$, number of iterations), such that the second-order methods compute better solutions compared to the standard value iteration scheme\footnote{The value of $w= 1.00001$ respects the constraint $w \leq w^*$ in all the three settings.}. For example, consider the $10$ states and $5$ actions setting (first row of Table \ref{table4}). The standard value iteration is run for $50$ iterations. It's per-iteration time is $0.2$ ms which results in an overall computational time of $0.0002 \times 50 = 0.01$ seconds. On the other hand, SOVI techniques (standard SOVI and the G-SOVI) are run for just $3$ iterations. However, their per-iteration time is $0.0033$ seconds and hence the overall computational time is $0.01$ seconds. We observe that, in $0.01$ seconds, the SOVI methods achieve lower error compared to the standard value iteration. Similarly, in the other two settings in Table \ref{table4}, we see that the second order SOVI algorithms achieve lower error compared to the standard value iteration when run for $0.02$ and $0.03$ seconds, respectively. 


It is important to note that this advantage need not hold for MDPs, in general, with large number of states and actions as the overhead for computing the Hessian inverse in large MDPs will be higher that would affect the overall computational time. If one could deploy techniques to improve the computation time for matrix operations, G-SOVI would be preferred for computing the optimal value function, over the standard value iteration.

%% file: conclusion.tex
\section{Conclusion}
In this work, we have proposed a generalized second-order value iteration scheme based on the Newton-Raphson method for faster convergence to near optimal value function in discounted reward MDP problems. The first step involved constructing a differentiable Bellman equation through an approximation of the $\max(.)$ operator. We then applied second order Newton method to arrive at the proposed algorithm. We proved the bounds on approximation error and showed second order convergence to the optimal value function. Finally, approaches geared towards easing the computational burden associated with solving problems involving large state and action spaces such as those based on approximate dynamic programming can be developed in the context of G-SOVI schemes in the future. 


%% file: main.bbl
\begin{thebibliography}{10}

\bibitem{bellman1966dynamic}
Richard Bellman.
\newblock Dynamic programming.
\newblock {\em Science}, 153(3731):34--37, 1966.

\bibitem{bertsekas1996neuro}
Dimitri~P Bertsekas and John~N Tsitsiklis.
\newblock {\em Neuro-dynamic programming}, volume~5.
\newblock Athena Scientific, Belmont, MA, 1996.

\bibitem{borkar2009stochastic}
Vivek~S Borkar.
\newblock {\em {Stochastic Approximation: A Dynamical Systems Viewpoint}}.
\newblock {Cambridge Univ. Press}, 2008.

\bibitem{dai2017sbeed}
Bo~Dai, Albert Shaw, Lihong Li, Lin Xiao, Niao He, Zhen Liu, Jianshu Chen, and
  Le~Song.
\newblock Sbeed: Convergent reinforcement learning with nonlinear function
  approximation.
\newblock {\em arXiv preprint arXiv:1712.10285}, 2017.

\bibitem{devraj2017zap}
Adithya~M Devraj and Sean Meyn.
\newblock Zap {Q-learning}.
\newblock In {\em Advances in Neural Information Processing Systems}, pages
  2235--2244, 2017.

\bibitem{furmston2016approximate}
Thomas Furmston, Guy Lever, and David Barber.
\newblock Approximate {Newton} methods for policy search in {Markov} decision
  processes.
\newblock {\em The Journal of Machine Learning Research}, 17(1):8055--8105,
  2016.

\bibitem{pymdp}
Github.
\newblock Python {MDPtoolbox}.
\newblock \url{https://github.com/sawcordwell/pymdptoolbox}.

\bibitem{goyal2019first}
Vineet Goyal and Julien Grand-Clement.
\newblock A first-order approach to accelerated value iteration.
\newblock {\em arXiv preprint arXiv:1905.09963}, 2019.

\bibitem{haarnoja2017reinforcement}
Tuomas Haarnoja, Haoran Tang, Pieter Abbeel, and Sergey Levine.
\newblock Reinforcement learning with deep energy-based policies.
\newblock In {\em Proceedings of the 34th International Conference on Machine
  Learning-Volume 70}, pages 1352--1361, 2017.

\bibitem{haarnoja2018soft}
Tuomas Haarnoja, Aurick Zhou, Pieter Abbeel, and Sergey Levine.
\newblock Soft actor-critic: Off-policy maximum entropy deep reinforcement
  learning with a stochastic actor.
\newblock {\em arXiv preprint arXiv:1801.01290}, 2018.

\bibitem{kamanchi2019successive}
Chandramouli Kamanchi, Raghuram~Bharadwaj Diddigi, and Shalabh Bhatnagar.
\newblock Successive over-relaxation {Q}-learning.
\newblock {\em IEEE Control Systems Letters}, 4(1):55--60, 2019.

\bibitem{lecun2015deep}
Yann LeCun, Yoshua Bengio, and Geoffrey Hinton.
\newblock Deep learning.
\newblock {\em Nature}, 521(7553):436, 2015.

\bibitem{mnih2013playing}
Volodymyr Mnih, Koray Kavukcuoglu, David Silver, Alex Graves, Ioannis
  Antonoglou, Daan Wierstra, and Martin Riedmiller.
\newblock Playing {Atari} with deep reinforcement learning.
\newblock {\em arXiv preprint arXiv:1312.5602}, 2013.

\bibitem{nesterov2005smooth}
Yu~Nesterov.
\newblock Smooth minimization of non-smooth functions.
\newblock {\em Mathematical programming}, 103(1):127--152, 2005.

\bibitem{ortega1970iterative}
James~M Ortega and Werner~C Rheinboldt.
\newblock {\em Iterative solution of nonlinear equations in several variables},
  volume~30.
\newblock SIAM, 1970.

\bibitem{puterman1979convergence}
Martin~L Puterman and Shelby~L Brumelle.
\newblock On the convergence of policy iteration in stationary dynamic
  programming.
\newblock {\em Mathematics of Operations Research}, 4(1):60--69, 1979.

\bibitem{reetz1973solution}
Dieter Reetz.
\newblock Solution of a {Markovian} decision problem by successive
  overrelaxation.
\newblock {\em Zeitschrift f{\"u}r Operations Research}, 17(1):29--32, 1973.

\bibitem{rust1994structural}
John Rust.
\newblock Structural estimation of {Markov} decision processes.
\newblock {\em Handbook of econometrics}, 4:3081--3143, 1994.

\bibitem{watkins1992q}
Christopher~J.C.H Watkins and Peter Dayan.
\newblock Q-learning.
\newblock {\em Machine learning}, 8(3-4):279--292, 1992.

\end{thebibliography}
